%% file: main.tex
\documentclass{article}

\makeatletter
\def\input@path{{iclr2027/}}
\makeatother

\usepackage{iclr2027_conference,times}

\input{math_commands.tex}


\usepackage{graphicx}
\usepackage{booktabs}
\usepackage{float}
\usepackage{xcolor}
\usepackage{tcolorbox}
\usepackage{amsthm}
\usepackage{amssymb}
\usepackage{algorithm}
\usepackage{algpseudocode}
\usepackage{wrapfig}
\usepackage{pifont}

\definecolor{tufte-red}{HTML}{990000}
\definecolor{tufte-green}{HTML}{2E6B3E}
\newcommand{\cmark}{\textcolor{tufte-green}{\ding{51}}}
\newcommand{\xmark}{\textcolor{tufte-red}{\ding{55}}}

\PassOptionsToPackage{hyphens}{url}
\usepackage[breaklinks=true,bookmarks=false]{hyperref}
\definecolor{lfm-blue}{HTML}{084A91}
\hypersetup{
    colorlinks=true,
    linkcolor=lfm-blue,
    citecolor=lfm-blue,
    urlcolor=lfm-blue
}
\usepackage{url}
\usepackage[capitalise,noabbrev]{cleveref}

\theoremstyle{definition}
\newtheorem{theorem}{Theorem}[section]

\newtheorem{proposition}[theorem]{Proposition}

\title{Langevin Flow Maps: Efficient Molecular \\ Dynamics and Transition Path Sampling}

\author{
  Sam McCallum$^{* \, 1 \,2}$, Niklas Rindtorff$^{*\, 2}$, Alexander Tong$^{2}$, James Foster$^{1}$ \\
  $^{*}$Equal contribution \\
  $^{1}$University of Bath \\
  $^{2}$AITHYRA
}

\iclrfinalcopy

\begin{document}

\maketitle

\begin{abstract}
    Molecular dynamics simulations proceed by integrating the Langevin equations over many small femtosecond timesteps. This poses a challenge for estimating ensemble properties and transition dynamics that occur on much longer timescales. We introduce Langevin Flow Maps, which extend machine-learned force-fields to additionally learn the stochastic Langevin integrator. We show that Langevin Flow Maps enable large-timestep molecular dynamics and recover accurate dynamical properties of the system, while running an order of magnitude faster than current machine-learned force fields. Further, by training on a diverse molecular dataset, we demonstrate a path towards \textit{transferable} Langevin Flow Maps.
\end{abstract}

\section{Introduction}
Molecular dynamics (MD) aims to compute the kinetic and equilibrium properties of a molecular system by numerically integrating equations of motion \citep{hollingsworth2018molecular}. These equations of motion are defined by a potential energy landscape, which in turn defines the forces acting on each atom in the system \citep{goldstein2011classical}.

In order to obtain accurate simulations we require that the potential energy landscape accurately describes the system's physics. Quantum Mechanical (QM) force calculations represent the most accurate computational method available for describing this landscape \citep{tuckerman2000understanding}. However, QM methods are significantly more expensive than classical approximations \citep{frenkel1997understanding}. This computational bottleneck has led to the development of machine-learned force fields (MLFFs) that attempt to learn the QM energy landscape with neural networks \citep{behler2007generalized, noe2020machine, keith2021combining, unke2021machine}. This field has shown great success at improving the efficiency of QM-accurate force calculations.

Nevertheless, force calculation represents one half of the puzzle; we must also integrate these forces over time \citep{fu2022forces}. This is commonly achieved by numerically integrating Langevin Dynamics \citep{leimkuhler2013rational}. To resolve the rapid oscillatory dynamics, timesteps of the order of 1 femtosecond (fs) are required. However, the transition and equilibrium dynamics of interest are typically observed beyond nanosecond timescales. Integration therefore remains a significant bottleneck in running molecular dynamics calculations \citep{wang2025design}.

As such, we may define the core problem for efficient molecular dynamics as to learn not only the instantaneous forces from data but to learn the integral of these forces too. Recent work has proposed solutions to this problem by learning transfer operators from a dataset of trajectories \citep{klein2023timewarp, schreiner2023implicit, diez2026transferable, thiemann2026force, bigi2026flashmd}. However, these methods require long molecular dynamics trajectories to train and their predictions are limited to the resolution of the saved trajectory.

Alternatively, flow map methods have emerged as a trajectory-free construction for learning the integral of a differential equation directly \citep{geng2026mean, boffi2026build}. Hamiltonian Flow Maps (HFMs) apply this machinery to learn the integral of Hamilton's equations of motion \citep{ripken2026learning}. Hamiltonian updates can be iterated alongside a stochastic update step such as the Langevin thermostat to sample equilibrium properties in molecular dynamics. Unfortunately, the Hamiltonian and stochastic updates do not commute and as a result this splitting scheme introduces a timestep-dependent bias in the transition dynamics; see \cref{subsec:background}.

In this work, we introduce Langevin Flow Maps (LFMs) that learn the stochastic solution map of Langevin Dynamics directly. LFMs enable large-timestep molecular dynamics to predict equilibrium properties \emph{and} transition dynamics of the system, without access to explicit trajectories as training data.

Our contributions can be summarized as follows:

\begin{enumerate}
    \item We construct a trajectory-free learning objective for the pathwise solution map of Langevin Dynamics.
    \item We demonstrate that LFMs allow for large-timestep molecular dynamics, predicting accurate equilibrium and dynamical properties.
    \item We train transferable LFMs on a diverse molecular dataset and show that the model is able to generalize to entirely unseen systems during inference.
\end{enumerate}

\section{Background \& Related Work}
\label{subsec:background}
Langevin Dynamics describes the dynamics of a particle subject to a potential energy landscape at constant temperature \citep{pastor1988analysis}. The system is described by the following coupled stochastic differential equation (SDE). For position $x_t$, velocity $v_t$ and potential $U(x_t)$, the system evolves via
\begin{equation}
    \label{eq:langevin-dynamics}
    \begin{aligned}
        dx_t &= v_t dt, \\
        dv_t &= \left[-\gamma v_t - m^{-1}\nabla U(x_t)\right] dt + \sigma dW_t,
    \end{aligned}
\end{equation}
where $m$ is the particle mass and $\gamma$ is the friction coefficient and $\sigma = \sqrt{2\gamma k_B T / m}$.

Langevin Dynamics admits a stationary distribution defined by the Hamiltonian $H(x, v) = U(x) + \frac{1}{2}m\lVert v \rVert^2$ with density $p \propto \exp(-\beta H(x, v))$, where $\beta = 1/k_B T$. In molecular dynamics this density corresponds to the canonical ensemble (NVT), denoted the Boltzmann distribution.

\paragraph{Numerical schemes.} To compute the solution to \cref{eq:langevin-dynamics}, numerical integration is used. The most widely used method in molecular dynamics is BAOAB~\citep{leimkuhler2013rational} that splits Langevin Dynamics into three pieces,
\begin{equation}
    \label{eq:baoab}
    \underbrace{dx_t = v_t dt}_{A},
    \qquad
    \underbrace{dv_t = -m^{-1}\nabla U(x_t) dt}_{B},
    \qquad
    \underbrace{dv_t = -\gamma v_t dt + \sigma dW_t}_{O}.
\end{equation}
The numerical approximation to each piece is then computed in the order in which the name suggests. Specifically, let $\mathcal{L}$ denote the generator of \cref{eq:langevin-dynamics}, then the solver approximates this transition via
\begin{equation}
    e^{\Delta t \mathcal{L}}
    = e^{\frac{\Delta t}{2}\mathcal{L}_B}
      e^{\frac{\Delta t}{2}\mathcal{L}_A}
      e^{\Delta t\mathcal{L}_O}
      e^{\frac{\Delta t}{2}\mathcal{L}_A}
      e^{\frac{\Delta t}{2}\mathcal{L}_B}
      + O(\Delta t^3).
\end{equation}
where $\mathcal{L}_N$ is the generator for step $N$. This splitting scheme incurs a local weak (distributional) error $O(\Delta t^3)$ and therefore a global weak error $O(\Delta t^2)$. Many small steps must therefore be taken to ensure accurate dynamics.

\paragraph{Hamiltonian Flow Maps.} Recent work by \citet{ripken2026learning} demonstrated how to learn the solution (or, flow) map of Hamilton's equations of motion. This corresponds to learning the solution map of the AB split from \cref{eq:baoab}. At inference, Hamiltonian Flow Maps iterate the learned AB map and the O step, 
\begin{equation}
    e^{\Delta t \mathcal{L}}
    = e^{\frac{\Delta t}{2}\mathcal{L}_O}
      e^{\Delta t\mathcal{L}_{AB}}
      e^{\frac{\Delta t}{2}\mathcal{L}_O}
      + O(\Delta t^3).
\end{equation}
This method suffers from the same fundamental non-commutativity of the chosen split as BAOAB and therefore the transition dynamics remain biased even for a perfectly learned Hamiltonian flow $e^{\Delta t\mathcal{L}_{AB}}$.

\paragraph{Transfer Operators} An alternative approach to simulating Langevin Dynamics has focused on learning the distributional transfer operator (TO), given by $p_{\Delta t}(x_{t+\Delta t} \mid x_t)$. These methods typically parameterize a generative model that approximates this conditional distribution \citep{klein2023timewarp, schreiner2023implicit, diez2026transferable, thiemann2026force, bigi2026flashmd}.

\begin{wraptable}[5]{r}{0.5\textwidth}
    \vspace{-1em}
    \centering
    \caption{Comparison of large-timestep molecular dynamics methods.}
    \label{tab:method-comparison}
    \small
    \begin{tabular}{lccc}
        \toprule
        Method & \shortstack{Equilibrium\\properties}
        & \shortstack{Trajectory\\free}
        & \shortstack{Transition\\dynamics} \\
        \midrule
        TOs & \cmark & \xmark & \cmark \\
        HFMs & \cmark & \cmark & \xmark \\
        \textbf{LFMs} & \cmark & \cmark & \cmark \\
        \bottomrule
    \end{tabular}
\end{wraptable}

The limitation of transfer operators is that they consume molecular dynamics trajectories as training data, requiring expensive classical integration to generate. Further, the learned timescales of the transfer operator are limited to the saved resolutions of the training data.

\section{Method}
The aim of this work is to learn the solution map of Langevin Dynamics directly, with transition operator $e^{\Delta t \mathcal{L}}$, that requires no explicit molecular dynamics trajectories as training data. To do this, we extend the method presented by \citet{mccallum2026strong} that demonstrates how to learn the pathwise solution map of an additive-noise SDE. The core idea is to note that the solution map of an SDE is a deterministic function given a realization of the Brownian path $W_{[s, t]} = \{W_u\}_{s \leq u \leq t}$. This enables an efficient regression objective without explicit trajectory supervision.

\subsection{Langevin Flow Maps}
\paragraph{Solution Map.} In the case of Langevin Dynamics, we can write the pathwise solution map as
\begin{equation}
    z_t = \Psi_{s, t}(z_s, W_{[s, t]}), \quad z_s = \begin{pmatrix}x_s \\ v_s\end{pmatrix}.
\end{equation}
That is, given an initial state $(x_s, v_s)$ and a Brownian path $W_{[s, t]}$, the final state $(x_t, v_t)$ is entirely determined.

This solution map satisfies a number of properties that can be exploited to parameterize the map for any molecular system. Notably, these properties can be satisfied given a dataset of molecular configurations and forces $(x^i, f^i)$ without access to molecular dynamics trajectories. It is therefore possible to recover the kinetic properties of a system without observing the dynamics directly.

\paragraph{Parameterization.} The solution map $\Psi_{s, t}$ can be expanded and written in terms of an integral equation,
\begin{equation}
\label{eq:uld-integral}
\begin{aligned}
    x_t &= x_s + \int_s^t v_u du, \\
    v_t &= v_s - \int_s^t \left( \gamma v_u - \frac{1}{m}f(x_u) \right)du + \sigma (W_t - W_s),
\end{aligned}
\end{equation}
where $f(x_t) = -\nabla U(x_t)$ and $\sigma = \sqrt{2\gamma k_B T / m}$. We therefore see that computing the solution map is equivalent to computing the velocity integral and the force integral. Additionally, since the stochastic process $(x_t, v_t)$ necessarily depends on the Brownian path, and we integrate directly over this process, the integrals themselves must depend on the Brownian path. We can therefore write the solution map as,
\begin{equation}
\label{eq:lfm}
\begin{aligned}
    x_t &= x_s + (t-s)v_{s, t}(z_s, W_{[s, t]}) \\
    v_t &= v_s - (t-s) \left[ \gamma v_{s, t}(z_s, W_{[s, t]}) - \frac{1}{m}f_{s, t}(z_s, W_{[s, t]}) \right] + \sigma (W_t - W_s),
\end{aligned}
\end{equation}
where the $(t-s)$ pre-factor is a modelling choice that forces $v_{s, t}$ and $f_{s, t}$ to represent the time-averaged integrals,
\begin{equation}
    v_{s, t}(z_s, W_{[s, t]}) = \frac{1}{t-s}\int_s^tv_u du, \quad f_{s, t}(z_s, W_{[s, t]}) = \frac{1}{t-s}\int_s^t f(x_u) du.
\end{equation}
This guarantees that as $s\rightarrow t$, the integrals $v_{s, t}$ and $f_{s, t}$ recover the instantaneous velocity $v_t$ and force $f(x_t)$, respectively.

\paragraph{Consistency conditions.} It is possible to show that the solution map to an SDE satisfies a set of consistency conditions~\citep{mccallum2026strong}. Specifically, the solution map $\Psi_{s, t}$ is uniquely determined if (1) it satisfies the correct instantaneous $s=t$ quantities and (2) it satisfies the semi-group property $\Psi_{s, t} = \Psi_{u, t} \circ \Psi_{s, u}$, where $u = (s+t)/2$.

We derive these conditions for Langevin Dynamics and demonstrate sufficiency in \cref{prop:consistency}. These consistency conditions will ultimately allow us to write an objective function which can be minimized to obtain the solution map (see \cref{subsec:training}).

\begin{tcolorbox}[colback=lfm-blue!6!white, colframe=lfm-blue!6!white, boxrule=0pt,
    arc=1mm, boxsep=0pt, left=8pt, right=8pt, top=2pt, bottom=2pt,
    before skip=6pt, after skip=6pt]
\begin{proposition}[Consistency condition]
    \label[proposition]{prop:consistency}
    Let $\Psi_{s, t}$ denote a candidate map to \cref{eq:uld-integral} parameterized by \cref{eq:lfm}. Then, for all $s, t \in \mathbb{R}$ and $u = (s+t)/2$, $\Psi_{s, t}$ is the solution map iff,
    \begin{equation*}
        v_{t, t}(z_t, W_{[t, t]}) = v_t, \quad f_{t, t}(z_t, W_{[t, t]}) = f(x_t),
    \end{equation*}
    and,
    \begin{equation*}
        \Psi_{s, t}(z_s, W_{[s, t]}) = \Psi_{u, t}(\Psi_{s, u}(z_s, W_{[s, u]}), W_{[u, t]}).
    \end{equation*}
\end{proposition}
\end{tcolorbox}

The proof of \cref{prop:consistency} is given in \cref{app:consistency}.

\subsection{Constructing the Brownian path}
\begin{wrapfigure}[19]{R}{0.5\textwidth}
  \vspace{-1.5em}
  \centering
  \includegraphics[width=\linewidth]{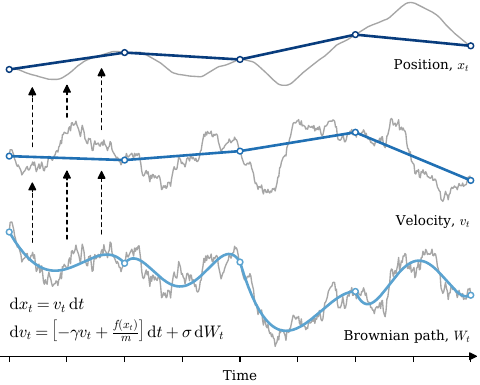}
  \caption{Langevin Flow Maps (\emph{blues}) learn the pathwise solution to Langevin Dynamics (\emph{gray}). Given a polynomial expansion of $W_{[s, t]}$, the flow map takes large timesteps to accurately predict the velocity and position trajectories.}
  \label{fig:overview}
\end{wrapfigure}

The solution map $\Psi_{s, t}$ depends on the Brownian path $W_{[s, t]}$ throughout the $[s, t]$ interval. We must therefore pass this information to the function in a way that captures the rich stochastic path features. \citet{mccallum2026strong} solve this by constructing a polynomial expansion to Brownian motion. The degree-4 polynomial expansion is shown in \cref{fig:overview}, where it can be seen to capture the path features over large time intervals.

This polynomial expansion has two essential properties: (1) its coefficients can be sampled exactly without first building the fine Brownian path and (2) polynomials can be composed over sub-intervals $[s, u], [u, t]$ into one polynomial over $[s, t]$. This second property enables an implementation of the semigroup condition. We discuss both of these properties below.

Firstly, the coefficients of this polynomial representation can be sampled independently as
\begin{equation}
    \label{eq:coeffs}
    I_{s, t}^{(n)} \sim \mathcal{N}\left(0, \frac{t-s}{2n + 1}\right).
\end{equation}
The zeroth coefficient, $I_{s, t}^{(0)} = W_t - W_s$, specifies the total increment; higher-order coefficients describe how the shape of the path departs from the straight line defined by this increment.

Secondly, the composition over sub-intervals is obtained via the Chen relations, denoted by
\begin{equation}
    I_{s, t}^{(n)} = I_{s, u}^{(n)} \oplus I_{u, t}^{(n)},
\end{equation}
where, for example, the zeroth coefficient combines as $I_{s, t}^{(0)} = I_{s, u}^{(0)} + I_{u, t}^{(0)}$. Higher order coefficients combine according to similar addition rules. Please see \cref{app:brownian} for the general expression.

\subsection{Training algorithm}
\label{subsec:training}
The consistency condition demonstrated by \cref{prop:consistency} enables an objective function which, when minimized, returns the solution map. We therefore look to parameterize the solution map as a neural network $\Psi_{s, t}^\theta$ and minimize this objective.

\paragraph{Neural parameterization.} Specifically, we implement the Langevin Flow Map as follows. For neural networks $v_{s, t}^\theta, f_{s, t}^\theta$ and polynomial coefficients $I_{s, t}^N = \{I_{s, t}^{(n)}\}_{0\leq n < N}$, we write
\begin{equation}
\label{eq:lfm-neural}
\begin{aligned}
    x_t &= x_s + (t-s)v_{s, t}^\theta(z_s, I_{s, t}^N) \\
    v_t &= v_s - (t-s) \left[ \gamma v_{s, t}^\theta(z_s, I_{s, t}^N) - \frac{1}{m}f_{s, t}^\theta(z_s, I_{s, t}^N) \right] + \sigma (W_t - W_s),
\end{aligned}
\end{equation}
where $W_t - W_s = I_{s, t}^{(0)}$ by definition of the zeroth coefficient. We define $\Psi_{s, t}^\theta := \Psi_{s, t}[v^\theta, f^\theta]$ to be the application of \cref{eq:lfm-neural} for a given $v^\theta, f^\theta$ tuple.

\paragraph{Objective function.} Translating the consistency condition into a training objective, we obtain
\begin{align}
    \mathcal{L}(\theta) &= \mathcal{L}_{v, f}(\theta) + \mathcal{L}_\text{D}(\theta),  \qquad \text{where} \\
    \mathcal{L}_{v, f}(\theta) &= \mathbb{E}_{t, v_t, x_t}\left[\lVert v_{t, t}^\theta(z_t, 0) - v_t\rVert^2 + \lVert f_{t, t}^\theta(z_t, 0) - f(x_t)\rVert^2 \right], \\
    \mathcal{L}_\text{D}(\theta) &= \mathbb{E}_{s, u, t, z_s, I}\left[\lVert \Psi_{s, t}^\theta(z_s, I_{s, t}^N) - \Psi_{u, t}^\theta(\Psi_{s, u}^\theta(z_s, I_{s, u}^N), I_{u, t}^N)\rVert^2 \right].
\end{align}

\begin{wrapfigure}[20]{r}{0.50\textwidth}
    \vspace{-2.0em}
    \begin{minipage}{\linewidth}
    \begin{algorithm}[H]
        \fontsize{8}{10}\selectfont
        \caption{Langevin Flow Map Training}\label{alg:lfm}
        \begin{algorithmic}[1]
            \Require Dataset $\mathcal{D}$, temperature $T$, batch size $M$ and split $\eta$
            \Statex \textbf{\# Sampling}
            \State $x^i, f(x^i) \sim \mathcal{D}, \quad v^i \sim \text{Maxwell-Boltzmann}(T)$
            \vspace{2pt}
            \Statex \textbf{\# Matching objective}
            \State Sample $(x^i, f(x^i), v^i)$ for $i = 0, \dots, \lfloor \eta M\rfloor$
            \State Compute loss
            \Statex $\mathcal{L}_{v, f}(\theta) = \mathbb{E}[\lVert v_0^\theta(z^i, 0) - v^i\rVert^2 + \lVert f_0^\theta(z^i, 0) - f(x^i)\rVert^2]$
            \par\vspace{2pt}
            \Statex \textbf{\# Semigroup objective}
            \State Sample $(x^i, f(x^i), v^i)$ for $i = 0, \dots, M - \lfloor \eta M\rfloor$
            \State Sample timesteps, $h^i \sim \log U(h_{\min}, h_{\max})$
            \State Sample $I_{s, u}^N, I_{u, t}^N \sim \mathcal{N}(0, h^i/(4n+2))$
            \State Chen combine $I_{s, t}^N = I_{s, u}^N \oplus I_{u, t}^N$
            \State Compute EMA teacher
            \Statex $\hat{z}_t = \texttt{stopgrad}[\Psi_{u, t}^{\hat{\theta}}(\Psi_{s, u}^{\hat{\theta}}(z^i, I_{s, u}^N), I_{u, t}^N)]$
            \State Compute student $z_t = \Psi_{h}^\theta(z^i, I_{s, t}^N)$
            \State Compute loss $\mathcal{L}_\text{D}(\theta) = \mathbb{E}\left[\lVert z_t - \hat{z}_t \rVert^2 / h^i \right]$
            \vspace{2pt}
            \Statex \# \textbf{Gradient}
            \State Update $\theta \leftarrow \theta - \lambda \nabla_\theta (\mathcal{L}_{v, f} + \mathcal{L}_\text{D})$
            \State EMA $\hat{\theta} \leftarrow \beta \hat{\theta} + (1-\beta)\theta$
        \end{algorithmic}
    \end{algorithm}
    \end{minipage}
\end{wrapfigure}

Note that at $s=t$ the coefficients are identically zero, $I_{t, t}^N = 0$; we therefore pass zero directly to the $s=t$ evaluations of the networks.

Evaluation of the training objective requires $[x, v, f(x)]$, plus a randomly sampled time-step $h=t-s$. Since Langevin Dynamics is a time-homogeneous system, no time-dependent data is required for training. This means that we can evaluate the objective with a dataset of uncorrelated-in-time positions and forces, $[x, f(x)]$, only. Correspondingly, the velocities, $v$, are sampled independently from the Maxwell-Boltzmann distribution at the desired temperature. 

\paragraph{Training algorithm.} The full training algorithm is presented in \cref{alg:lfm}. For the semigroup loss we use the self-distillation formulation, where the two-step map is implemented using an exponential moving average (EMA) of the network's weights. This EMA network is called the teacher. We apply $\texttt{stopgrad}$ to the output of the two-step teacher following common practice \citep{boffi2026build, mccallum2026strong}. These two choices have been shown to stabilize training with self-distillation objectives \citep{frans2025one}.

\section{Experiments}
The experiments are ordered in two sections: single-molecule dynamics and transferable dynamics. The single-molecule section trains an LFM to learn the forces and solution map of each molecule individually, where we use the MD17 small-molecules dataset from \citet{chmiela2017machine} and the alanine dipeptide dataset from \citet{kohler2021smooth}. The transferable section applies one LFM model to a large dataset of peptide sequences from \citet{tan2025manypeptides} and investigates the integration performance when generalizing to unseen peptides.

\subsection{Single-molecule Dynamics}
\subsubsection{Organic molecules}
We first investigate the performance of the LFM for large-timestep molecular dynamics when trained on single-molecule datasets. We use the MD17 dataset, which releases equilibrium configurations and forces for a set of organic molecules computed via density functional theory \citep{chmiela2017machine}.

\begin{figure}
  \centering
  \includegraphics[width=\textwidth]{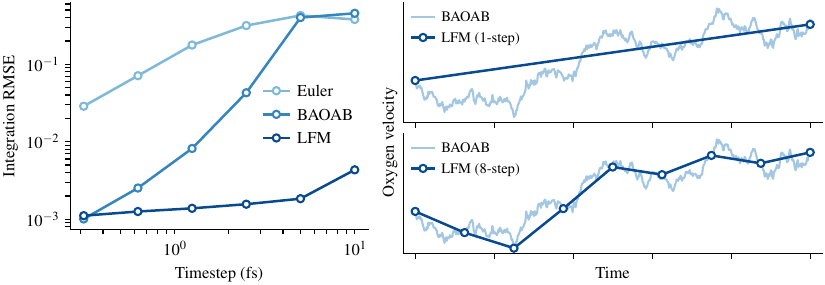}
  \caption{\emph{Left}.\ Integration RMSE for the learned ethanol dynamics. \emph{Right}.\ Pathwise solution comparison between the fine reference BAOAB trajectory and the large-timestep LFM prediction.}
  \label{fig:ethanol-strong-error}
\end{figure}

\paragraph{Integration Error.} To measure the accuracy of the pathwise solution, we compute the integration root-mean-squared-error (RMSE) between the predicted solution and a fine-timestep reference. The results for the ethanol molecule are shown in \cref{fig:ethanol-strong-error}. We see that the LFM at $10$fs timesteps obtains comparable accuracy to BAOAB at $1$fs, while also improving upon the small-timestep accuracy.

\begin{figure}
  \centering
  \includegraphics[width=\textwidth]{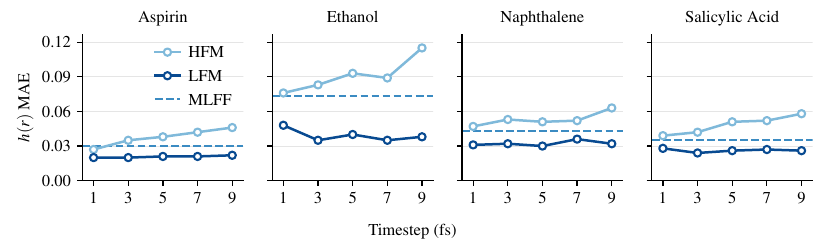}
  \caption{Interatomic distances $h(r)$ MAE as a function of step size for 300 ps of MD simulation. Reference MLFF shown with $0.5$fs timestep integration.}
  \label{fig:interatomic}
\end{figure}

\paragraph{Molecule Structure.} Next we compare the molecular structure predicted by the LFM as a function of the timestep. The distribution of interatomic distances, $h(r)$, is calculated for the predicted molecules and compared to the reference dataset. The LFM is compared to Hamiltonian Flow Maps and the reference MLFF baseline \citep{ripken2026learning}; see \cref{fig:interatomic}.

We see that the MAE obtained by the LFM predictions is lower than that obtained by the HFM for all timesteps and molecules. Additionally, the LFM predictions achieve a lower MAE than the reference MLFF at $0.5$fs timesteps. This demonstrates that the LFM has not only learned an accurate large-timestep integrator, but is able to improve upon the integration error at the small time steps commonly used to run molecular dynamics.

\begin{wraptable}[8]{r}{0.45\textwidth}
    \vspace{-2.0em}
    \centering
    \caption{Force MAE in units of meV/\AA.}
    \vspace{1pt}
    \label{tab:md17-force-mae}
    \begin{tabular}{lc|cc}
      \toprule
      Molecule & MLFF & HFM & LFM \\
      \midrule
      Aspirin        & \textbf{3.8} & 7.0 & 5.9 \\
      Ethanol        & \textbf{1.5} & 2.8 & 2.9 \\
      Naphthalene    & \textbf{1.5} & 3.0 & 2.8 \\
      Salicylic acid & \textbf{2.2} & 4.7 & 4.3 \\
      \bottomrule
    \end{tabular}
\end{wraptable}

\paragraph{Force Error.} The LFM implements the standard force matching loss to constrain the $h=0$ force network. We can therefore compare the force accuracy obtained by the LFM to standard MLFF training. This is shown in \cref{tab:md17-force-mae}. We see that the force error obtained by the MLFF is lower than the LFM, as expected. However, when analyzed alongside \cref{fig:interatomic} where the LFM obtains a higher structural accuracy than the MLFF, it becomes clear that force error alone is not a sufficient training objective for accurate molecular dynamics. This result verifies the sentiment developing in the MLFF literature that `forces are not enough' \citep{fu2022forces}.

\begin{figure}
  \centering
  \includegraphics[width=\textwidth]{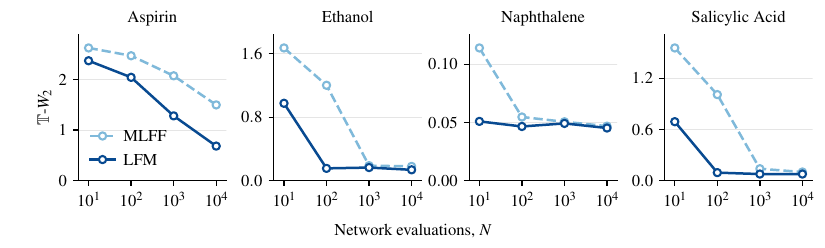}
  \caption{Convergence to the Boltzmann distribution measured by the 2-Wasserstein distance over torsion angles, $\mathbb{T}$-$W_2$, as a function of the number of network evaluations, $N$.}
  \label{fig:md17-boltzmann-convergence}
\end{figure}

\paragraph{Boltzmann Convergence.} A key property of Langevin Dynamics is that the sample distribution converges to the Boltzmann distribution. This convergence occurs over long timescales and therefore many function evaluations of the force-field are required. To compare the same infinitesimal dynamics, we define the MLFF by the LFM's force network evaluated at $h=0$. We compare the Boltzmann convergence speed of the MLFF integrated with BAOAB at $0.5$fs and the LFM at $9$fs in terms of the number of function evaluations. We measure convergence by the 2-Wasserstein distance on the torsional molecule angles from a biased initialization; see \cref{fig:md17-boltzmann-convergence}. We see that the convergence speed of the LFM is faster than the integrated MLFF. For example, the ethanol / naphthalene / salicylic acid distributions predicted by the LFM have converged after 100 function evaluations compared to the 1,000 evaluations required by the MLFF.

\subsubsection{Alanine Dipeptide}
We next train the LFM on the canonical alanine dipeptide system which admits multiple meta-stable torsional states with slow mixing times \citep{kohler2021smooth}. We use this system to investigate the LFMs ability to recover accurate dynamical properties and sample transition pathways.

\begin{figure}
  \centering
  \includegraphics[width=\textwidth]{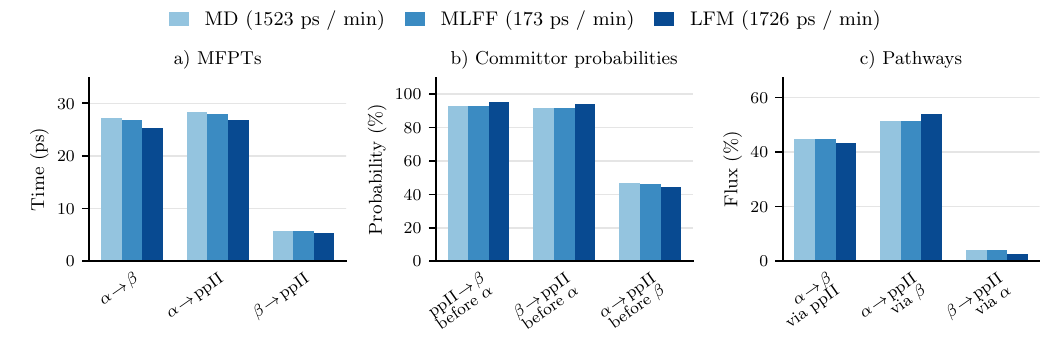}
  \caption{Alanine dipeptide transition dynamics for trajectories generated by each method. Integration throughput is shown in picoseconds per minute, measured on 1x RTX A6000.}
  \label{fig:msm-aldp}
\end{figure}

\paragraph{Transition dynamics.} In \cref{fig:msm-aldp}, we show the transition dynamics between meta-stable states predicted by the LFM alongside the MLFF and reference MD trajectory. We see that the transition dynamics obtained by the LFM are comparable to the MLFF, while performing the calculation significantly faster. Additionally, the LFM predicts transition metrics accurate to classical MD using the Amber99SB force-field and even improves upon the integration speed. This indicates the possibility for obtaining QM-accurate molecular dynamics at comparable speed to classical mechanics.

\begin{figure}
  \centering
  \includegraphics[width=\textwidth]{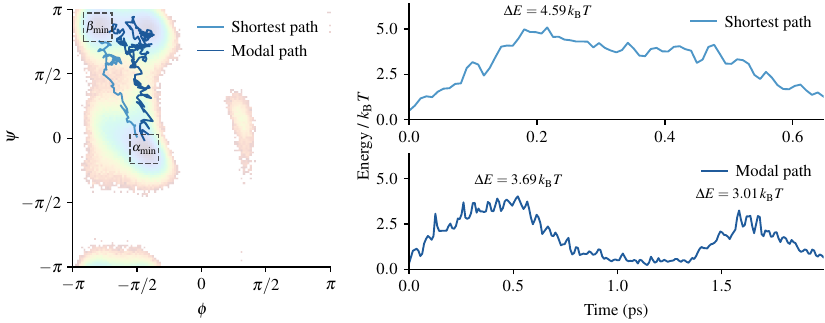}
  \caption{Transition paths for the $\alpha \rightarrow \beta$ transition in alanine dipeptide. Activation energies are shown for each free energy maximum, measured from the preceding free energy minima.}
  \label{fig:transition-paths}
\end{figure}

\paragraph{Pathwise observables.} In \cref{fig:transition-paths}, the transition paths between the $\alpha$ and $\beta$ states are shown alongside the free energy observed during the transition. Specifically, a batch of molecules are seeded in the $\alpha$ state and integrated with the LFM until reaching the $\beta$ state. This enables the transition time density to be calculated. Then, the shortest path and modal path are extracted from this distribution. Interestingly, we see that the shortest path walks the transition by directly overcoming one large activation energy barrier whereas the modal path transitions via polyproline (ppII) and overcomes two smaller activation energy barriers.

\subsection{Transferable Dynamics}
The LFM architecture requires only atomic numbers to define the molecular system and therefore naturally extends to modelling a diverse molecular input space. We use this property to train a transferable LFM on a diverse set of peptide conformations from the Many Peptides dataset \citep{tan2025manypeptides} and investigate the extent to which the LFM can generalize to accurately integrate unseen peptide sequences. We investigate the integration error, Boltzmann convergence and transition dynamics predicted by the model for held-out peptide sequences. See \cref{app:experiments} for details.

\begin{wrapfigure}[14]{r}{0.40\textwidth}
  \vspace{-1em}
  \centering
  \includegraphics[width=\linewidth]{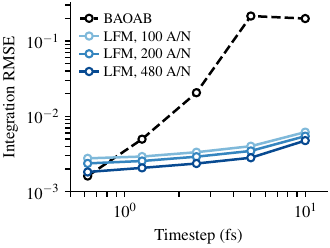}
  \caption{Transferable LFM integration error on \textit{Many Peptides}.}
  \label{fig:transfer-strong-timestep}
\end{wrapfigure}

\paragraph{Integration Error.} In \cref{fig:transfer-strong-timestep}, we calculate the integration RMSE of the transferable LFM on 92 unseen peptide sequences, ranging from dipeptides through to decapeptides. We compute the error throughout training as measured by the number of atoms per parameter (A/N). We see that at $1$fs timesteps the LFM obtains comparable integration error to BAOAB, while maintaining a significantly lower error up to $10$fs timesteps.

\paragraph{Boltzmann convergence.} In \cref{tab:lfm-timestep-ensemble}, we investigate the LFMs ability to recover the Boltzmann distribution when initialized from a single conformation of an unseen peptide sequence. We integrate the peptide for $10^7$ function evaluations and report the cumulative energy and backbone torsion 2-Wasserstein distances ($\mathcal{E}$-$W_2$, $\mathbb{T}$-$W_2$). We see that the ensemble metrics improve as the LFMs timestep decreases, with BAOAB integration on the LFMs $h=0$ force network returning metrics comparable to the ground-truth Amber14 force-field.

\paragraph{Transition dynamics.} Additionally, in \cref{tab:lfm-timestep-ensemble}, we calculate the meta-stable state occupation, identified by a time-lagged independent component analysis (TICA), and compute transition probabilities between these states. We compute the total variation over the state occupation $\pi$-TV and transition $\mathbb{P}$-TV densities with respect to the reference dataset. The large-timestep LFM metrics consistently outperform the Amber14 force-field, obtaining more accurate meta-stable state occupation and transition probabilities.

\paragraph{State discovery.} We see from \cref{tab:lfm-timestep-ensemble} that the small-timestep dynamics more accurately represent the Boltzmann ensemble whereas the large-timestep dynamics best represent the meta-stable state occupation and transition dynamics. This presents the opportunity to initially explore the state-space with large LFM timesteps and perform a final relaxation with small-timestep integration. We demonstrate the advantage of this procedure in \cref{fig:octa-exploration}, where the LFM discovers a meta-stable mode not reached by long molecular dynamics rollouts.

\begin{table}
    \centering
    \caption{Transferable LFM sampling on nine unseen peptide sequences with length 2, 4, 8 from \textit{Many Peptides}. Each metric is reported as the mean across three peptides at each length.}
    \label{tab:lfm-timestep-ensemble}
    \input{figures/sampling_table.tex}
\end{table}

\begin{figure}
    \centering
    \includegraphics[width=\linewidth]{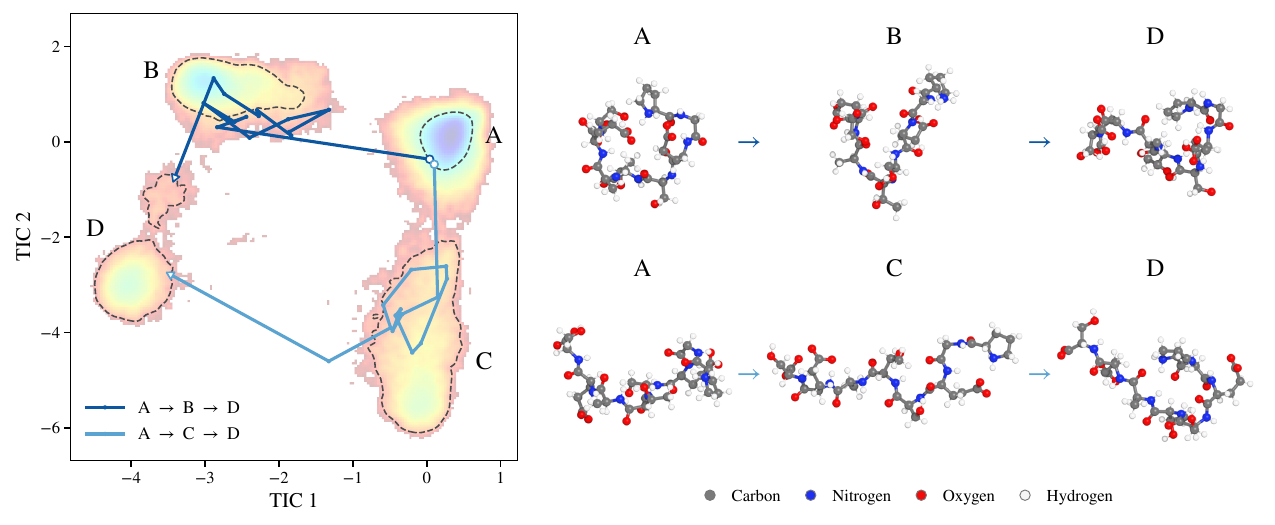}
    \caption{\emph{Left}.\ Transition pathways for octapeptide PGESTAES predicted by the LFM and projected onto the first two time-lagged independent components (TIC). \emph{Right}.\ Peptide conformations measured along each transition pathway.}
    \label{fig:octa-tps}
\end{figure}

\paragraph{Transition pathways.} We can apply the transferable LFM to calculate transition paths between meta-stable states defined by the TICA. In \cref{fig:octa-tps}, we show the free energy surface for the PGESTAES octapeptide and the four meta-stable states obtained via clustering. To obtain transition pathways, the conformer is seeded in state A and integrated with the LFM at $4$fs until reaching state D. We plot two coarsened pathways that transition via states B and C with the corresponding molecular conformations along each path. This peptide sequence was not seen by the LFM during training, yet we see that the model is able to recover consistent transition pathways.

\section{Conclusions} In this work, we have introduced a method to learn the solution map of Langevin dynamics, termed Langevin Flow Maps (LFMs). This construction extends machine-learned force fields to learn both the instantaneous forces and the integrated system dynamics directly without access to explicit trajectory data. We demonstrated that LFMs enable large-timestep molecular dynamics while predicting accurate ensemble properties and transition dynamics. Further, a transferable LFM model trained on a diverse molecular dataset was shown to generalize to unseen peptide sequences, indicating the possibility of a foundational molecular dynamics integrator.

\subsection*{Author contributions}
SM and NR conceived the idea and jointly developed the theory.
SM led the single-system LFM experiments and NR led the transferable LFM experiments. SM and NR both developed the open-source training code. SM and NR jointly wrote the paper, with contributions from all other authors.
AT and JF guided the project.

\subsection*{AI use statement}
In this work, we used generative AI tools to assist with writing code to run the experiments, searching the literature for related work and text editing to improve clarity. All code, literature and text edits were checked manually for correctness. We take responsibility for the final content of this work, including text, claims and artifacts produced.

\subsection*{Ethics statement}
This work contributes to the field of molecular dynamics by improving simulation efficiency. The method may support applications in chemistry, materials science and drug design. We do not anticipate additional ethical concerns from this work beyond those established in the molecular dynamics literature.

\subsection*{Reproducibility statement}
The LFM training code is available on GitHub at \href{https://github.com/sammccallum/lfm}{sammccallum/lfm}.
The Many Peptides Forces dataset is available on Hugging Face at \href{https://huggingface.co/datasets/niklastr/many_peptides_forces}{niklastr/many\_peptides\_forces}.
Model checkpoints, rollout trajectories, and reference trajectories are available on Hugging Face at \href{https://huggingface.co/niklastr/langevin_flow_maps}{niklastr/langevin\_flow\_maps}.
Model architectures, training settings and evaluation procedures are described in the appendix.

\bibliography{references}
\bibliographystyle{iclr2027/iclr2027_conference}

\newpage
\appendix
\crefalias{section}{appendix}
\crefalias{subsection}{appendix}
\crefalias{subsubsection}{appendix}
\section{Theory}
\subsection{Brownian Motion}
\label{app:brownian}

First introduced by \citet{foster2020optimal} and \citet{habermann2021semicircle}, the degree-N polynomial expansion of Brownian motion over $[s, t]$ is given by,
\begin{equation}
    \label{eq:polyapprox}
    W_{u,v}^{(N)} = \sum_{n=0}^{N-1} \frac{2n+1}{t-s} I_{s,t}^{(n)} \int_u^v \widetilde{P}_n\left(\frac{r-s}{t-s}\right) d r,
\end{equation}
where $\widetilde{P}_n$ is the $n$-th shifted Legendre polynomial on $[0, 1]$ and $I_{s, t}^{(n)}$ are the polynomial coefficients given by,
\begin{equation}
    \label{eq:coeffs-integral}
    I_{s,t}^{(n)} = \int_s^t  \widetilde{P}_n\left(\frac{u-s}{t-s}\right) d W_u.
\end{equation}

For example, taking the degree-3 representation we get
\begin{equation}
    W_{s,v}^{(3)}
    = I_{s,t}^{(0)}x
    + 3I_{s,t}^{(1)}(x^2-x)
    + 5I_{s,t}^{(2)}(2x^3-3x^2+x),
    \qquad x=\frac{v-s}{t-s}.
\end{equation}

\paragraph{Properties of $I_{s, t}$.} Since the coefficients of the polynomial expansion are defined by the stochastic integral in \cref{eq:coeffs-integral}, they are normally distributed and can be sampled directly \citep{mccallum2026strong}. Specifically, the coefficients are distributed by
\begin{equation}
    I_{s, t}^{(n)} \sim \mathcal{N}\left(0, \frac{t-s}{2n + 1}\right).
\end{equation}
Additionally, these coefficients can be combined over sub-intervals, $[s, u], [u, t]$, to form the coefficients over the full interval, $[s, t]$. This is achieved by the following Chen relations,
\begin{equation}
\label{eq:polychen}
    I_{s,t}^{(n)}=\sum_{m=0}^n c_{n,m}\Big (I_{s,u}^{(m)}+(-1)^{n+m}I_{u,t}^{(m)}\Big ),
\end{equation}
where,
\begin{equation}
    c_{n,m} :=(-1)^{n}(2m+1)\sum_{k=m}^n\Big (-\frac 12\Big )^k \frac{(n+k)!}{(n-k)!(k-m)!(k+m+1)!}.
\end{equation}
For example, the $n=0$ and $n=1$ coefficients combine as
\begin{align}
I_{s,t}^{(0)}
&= I_{s,u}^{(0)} + I_{u,t}^{(0)}, \\
I_{s,t}^{(1)}
&= \frac12\left(I_{s,u}^{(1)} + I_{u,t}^{(1)}\right)
 + \frac12\left(I_{u,t}^{(0)} - I_{s,u}^{(0)}\right).
\end{align}

\subsection{Consistency condition}
\label{app:consistency}
In this section we provide a short proof of \cref{prop:consistency}. This follows directly from \citet[Proposition 3.3]{mccallum2026strong}. The statement is reproduced here for convenience. We inherit the standard regularity assumptions on the drift and diffusion fields \citep[Assumption A.1]{mccallum2026strong}.

\begin{proposition}[{\citet[Proposition 3.3]{mccallum2026strong}}]
\label[proposition]{prop:mccallum}
Let $\Psi_{s, t}(z_s, W_{[s, t]})$ denote the solution map of the following additive-noise SDE,
\begin{equation}
    \label{eq:additive-sde}
    dz_t = b(t, z_t) dt + g(t) dW_t,
\end{equation}
which be written as 
\begin{equation}
    \Psi_{s, t}(z_s, W_{[s, t]}) = z_s + b_{s, t}(z_s, W_{[s, t]}) (t-s) + g_{s, t}(W_{[s, t]}) (W_t - W_s).
\end{equation}
Then $\Psi_{s, t}$ is the solution map if and only if
\begin{align}
    b_{t,t}(z_t) &= b(t,z_t), \quad g_{t,t} = g(t), \quad \text{and}
    \label{eq:matching-mccallum} \\
    \Psi_{s,t}(z_s,W_{[s,t]})
    &= \Psi_{u,t}(\Psi_{s,u}(z_s,W_{[s,u]}),W_{[u,t]}).
    \label{eq:semi-group-mccallum}
\end{align}
\end{proposition}

To show sufficiency of \cref{prop:consistency} for obtaining the solution map to Langevin Dynamics, it is therefore sufficient to demonstrate the equivalence to \cref{prop:mccallum}. First, consider that the Langevin Dynamics SDE can be written in the form of \cref{eq:additive-sde} as
\begin{equation}
    d \begin{bmatrix}
        x_t \\ v_t
    \end{bmatrix} = \begin{bmatrix}
        v_t \\
        -\gamma v_t + m^{-1}f(x_t)
    \end{bmatrix} dt + \begin{bmatrix}
        0 \\ \sigma
    \end{bmatrix} dW_t.
\end{equation}
Then, the matching condition in \cref{eq:matching-mccallum} requires
\begin{equation}
    b_{t, t}(z_t) = \begin{bmatrix}
        v_t \\
        -\gamma v_t + m^{-1}f(x_t)
    \end{bmatrix},
\end{equation}
which is equivalent to $v_{t, t}(z_t, W_{[t, t]}) = v_t$ and $f_{t, t}(z_t, W_{[t, t]}) = f(x_t)$ with $W_{[t, t]} := \varnothing$.

Finally, the semigroup condition \cref{eq:semi-group-mccallum} required by \cref{prop:mccallum} is retained exactly in \cref{prop:consistency}. This completes the proof.

\section{Model Architecture}
\label{app:architecture}
Following the Hamiltonian Flow Map architecture of \citet{ripken2026learning}, we parameterize $v_{s, t}^\theta$ and $f_{s, t}^\theta$ using a shared translation-invariant transformer. The architectural addition required to learn the Langevin Flow Map is to pass the Brownian polynomial coefficients $I_{s, t}^N$ as conditioning information. We describe the full architecture here for completeness.

\paragraph{Embeddings.} For atom $i$ with atomic number $Z_i$, we initialize a token $h_i^{(0)}$ and construct relative positional embeddings,
\begin{equation}
    h_i^{(0)} = \operatorname{Embed}(Z_i), \qquad
    e_{ij} = \operatorname{MLP}\left([\phi(d_{ij}); \hat{r}_{ij}]\right),
\end{equation}
where $d_{ij}=\lVert x_i-x_j\rVert$, $\hat{r}_{ij}=(x_i-x_j)/d_{ij}$ and $\phi$ is a cosine basis expansion. Here $x_i$ denotes the initial position and $[\cdot;\cdot]$ denotes concatenation.

\paragraph{Attention.} For a single attention head, the positional embeddings enter the attention operation via
\begin{equation}
\begin{aligned}
    a_{ij} &= \frac{(W_Q h_i)^\top (W_K h_j \odot W_{K,e}e_{ij})}{\sqrt{d_h}}, \\
    \operatorname{Attn}(h)_i &= \sum_j
    \operatorname{softmax}_j(a_{ij})
    \left(W_V h_j \odot W_{V,e}e_{ij}\right),
\end{aligned}
\end{equation}
where $\odot$ denotes element-wise multiplication.

\paragraph{Conditioning.} The time-step $h=t-s$, initial velocity $v_i$ and mass $m_i$ are embedded as
\begin{equation}
    c^h = \operatorname{MLP}_h(\phi_h(h)), \quad
    c_i^m = \operatorname{MLP}_m(m_i), \quad
    c_i^v = \operatorname{MLP}_v\left([W_{Z_i}\psi_v(\lVert v_i\rVert); v_i]\right),
\end{equation}
where $\phi_h$ denotes Gaussian Fourier features with learnable frequencies, $\psi_v$ is a Gaussian basis expansion and $W_{Z_i}$ is an element-specific matrix. We add a Brownian embedding using the coefficient vectors $I_{s,t,i}^{(n)}\in\mathbb{R}^3$ associated with atom $i$ via
\begin{equation}
    c_i^I = \operatorname{MLP}_I\left(\left[\psi_I\left(\lVert I_{s,t,i}^{(n)}\rVert\right); I_{s,t,i}^{(n)}\right]\right),
\end{equation}
and obtain the total conditioning input with
\begin{equation}
    c_i = \operatorname{MLP}\left([c^h; c_i^v; c_i^I; c_i^m]\right).
\end{equation}
Each transformer block uses $c_i$ to compute separate scale, shift and gating parameters for the hidden state, given by
\begin{equation}
\begin{aligned}
    \left[\gamma_i;\beta_i;\alpha_i\right] &= W\operatorname{SiLU}(\operatorname{LN}(c_i))+b, \\
    \operatorname{AdaLN}(h_i;\gamma_i,\beta_i) &= (1+\gamma_i)\odot\operatorname{LN}(h_i)+\beta_i, \\
    h_i &\leftarrow h_i + \alpha_i\odot\operatorname{Block}\left(\operatorname{AdaLN}(h;\gamma,\beta)\right)_i,
\end{aligned}
\end{equation}
where $\operatorname{Block}$ denotes the attention or feed-forward update.

\paragraph{Prediction heads.} After $L$ transformer blocks, separate heads predict the mean velocity and force,
\begin{equation}
\begin{aligned}
    v_{s,t,i}^\theta &= \operatorname{MLP}_v^{\mathrm{out}}\left(\operatorname{AdaLN}(h_i^{(L)};\gamma_i^v,\beta_i^v)\right), \\
    f_{s,t,i}^\theta &= \operatorname{MLP}_f^{\mathrm{out}}\left(\operatorname{AdaLN}(h_i^{(L)};\gamma_i^f,\beta_i^f)\right).
\end{aligned}
\end{equation}
The task-specific scale and shift parameters are obtained from $c_i$ as above. These predictions define the Langevin Flow Map through \cref{eq:lfm-neural}. The hyperparameters used for this architecture are given in \cref{tab:architecture}.

\begin{table}[H]
    \centering
    \caption{Model hyperparameters.}
    \label{tab:architecture}
    \begin{tabular}{lc}
        \toprule
        Hyperparameter & Value \\
        \midrule
        Transformer blocks & 6 \\
        Attention heads & 8 \\
        Hidden dimension & 256 \\
        Distance basis functions & 10 \\
        Velocity Gaussian basis functions & 8 \\
        Brownian Gaussian basis functions & 8 \\
        Brownian coefficients & 4 \\
        MLP layers & 2 \\
        Feed-forward hidden width & 1024 \\
        \bottomrule
    \end{tabular}
\end{table}

\section{Experimental Details}
\label{app:experiments}
\subsection{Single-molecule Dynamics}
We train an LFM for each MD17 molecule at $500$\,K with friction $\gamma=10$\,ps$^{-1}$. The model hyperparameters are shown in \cref{tab:architecture}. The dataset is split into training, validation and test sets in an $80{:}10{:}10$ ratio, with the same split as \cite{ripken2026learning}.

During training, each configuration is centered and a random rotation is applied. The models are trained for $100{,}000$ updates with batch size 512 using Adam ($\beta_1=0.9$, $\beta_2=0.95$) and learning rate warmup from $10^{-6}$ to $10^{-3}$ over 2,000 steps before cosine decay to a final learning rate of $10^{-7}$. The fraction of the batch allocated to the matching objective is $\eta=0.25$ and we use EMA decay $0.99$, gradient clipping at norm 10 and log-uniform timesteps between $0.01$ and $10$fs.

\paragraph{Molecule Structure.} We initialize 256 trajectories from test configurations with thermal velocities and integrate for 300ps. Following \citet{ripken2026learning}, the interatomic distance distribution for a configuration $x$ of $N$ atoms is
\begin{equation}
    h(r) = \frac{1}{N(N-1)}\sum_{i=1}^{N}\sum_{\substack{j=1 \\ j\neq i}}^{N}
    \delta\!\left(r-\lVert x_i-x_j\rVert_2\right),
\end{equation}
where $\delta$ is the Dirac delta.

The MAE between the generated distribution, $h(r)$, and the reference dataset distribution $\hat{h}(r)$ is calculated as
\begin{equation}
    \operatorname{MAE}_{h(r)} = \int_0^\infty
    \left|\langle h(r)\rangle-\langle\hat{h}(r)\rangle\right|\,dr,
\end{equation}
where $\langle \cdot \rangle$ denotes the mean. Frames from the first bond-length deviation exceeding $0.5$\,\AA\ from the mean are excluded from each trajectory \citep{ripken2026learning, fu2022forces}.

\paragraph{Boltzmann Convergence.} We initialize 256 trajectories at the same configuration with independently sampled velocities and Brownian coefficients. Integration is performed for $N=10{,}000$ network evaluations with both the LFM and BAOAB. At each evaluation, we compare the torsion-angles of the ensemble to 256 configurations sampled from the reference dataset using the periodic 2-Wasserstein distance.

\subsection{Alanine Dipeptide}
We use the same data split, architecture and training settings as for the single-molecule experiments, but with a temperature of $300$\,K and $200{,}000$ training steps.

\paragraph{Transition dynamics.} We initialize 256 trajectories and integrate for $1$ns with the LFM at $10$fs and BAOAB with the LFM's $h=0$ force network at $1$fs. The MD reference is a $1\mu$s OpenMM trajectory after $1$ns equilibration, using Amber99SB-ILDN with OBC implicit solvent and a $1$fs timestep. Frames are saved every $1$ps.

Each trajectory is fit with a maximum-likelihood Markov State Model (MSM) using a $1$\,ps lag. The discrete states of the MSM are chosen to be the $\alpha, \beta$, ppII meta-stable states with an additional state to handle the remaining state-space. The backbone angles $(\phi,\psi)$ of each state are defined by: $\alpha$ with $\phi\in[-160,-20]^\circ$ and $\psi\in[-120,50]^\circ$; $\beta$ with $\phi\in[160,270]^\circ$ and $\psi\in[50,240]^\circ$; ppII with $\phi\in[-90,-20]^\circ$ and $\psi\in[50,240]^\circ$.

\paragraph{Pathwise observables.} We initialize 256 trajectories from the $\alpha$ state and integrate with the LFM at $9$fs timesteps until first entry into the $\beta$ state. We show the shortest path and the path nearest the mode of the transition time density. The plotted free energy is obtained from the reference-data histogram.

\subsection{Transferable Dynamics}
\subsubsection{Training \& Inference}
We use \textit{Many Peptides Forces}, which augments the ManyPeptidesMD training collection \citep{tan2025manypeptides} with energy and force labels for 434M configurations of $21{,}700$ peptides of length 2--8. The 9M parameter LFM model was trained with hyperparameters from \cref{tab:architecture} at three training compute budgets, expressed in atoms per parameters (100, 200, and 480).

We use the same single-molecule architecture and training setup described above, but supply input atomic numbers and masses for each molecular configuration. To batch systems with different numbers of atoms, we pad the atom arrays to a maximum size and mask padded atoms throughout the transformer and flow map forward pass. At inference, we additionally apply a temperature corrector algorithm to stabilize the kinetic temperature, described below. 

\paragraph{Temperature corrector.} The learned flow map velocity and force heads are not perfectly trained to zero loss. The remaining bias enters each update and over many steps can introduce a bias to the kinetic temperature. This can be corrected by making small changes to the Langevin thermostat at each step.

For a group of atoms with mass $m$, let $v_i$ denote the predicted velocities and $n_i=\sigma I_{s,t,i}^{(0)}$ the explicit noise updates. We look to obtain corrected velocities $\hat{v}_i=v_i+\alpha n_i$ satisfying
\begin{equation}
    \label{eq:temp-constraint}
 \mathbb{E}[\lVert\hat{v}_i\rVert^2]=R, \qquad R=3k_BT/m.
\end{equation}

Then, defining
\begin{equation}
    A=\mathbb{E}[\lVert v_i\rVert^2], \qquad
    B=\mathbb{E}[v_i\cdot n_i], \qquad
    C=\mathbb{E}[\lVert n_i\rVert^2],
\end{equation}
and expanding \cref{eq:temp-constraint}, we get
\begin{equation}
    C\alpha^2+2B\alpha+A-R=0.
\end{equation}
This quadratic can be solved to obtain a real-valued $\alpha$ that corrects velocities to the target temperature $T$. Since the corrected noise update is given by $(1 + \alpha) \sigma I_{s, t, i}^{(0)}$ with $1 + \alpha \geq 0$, this update simply scales the noise magnitude and preserves the Langevin update. If no permissible root exists, then we fallback to a velocity rescaling $\hat{v}_i = \sqrt{R/A}v_i$.

\subsubsection{Metrics}
\paragraph{Ensemble metrics.}
We evaluate the cumulative distribution of configurations from $10^7$ integration steps against the reference dataset. We compute the energy $\mathcal{E}$-$W_2$ and backbone torsion $\mathbb{T}$-$W_2$ 2-Wasserstein distance metrics, following \citet{tan2025manypeptides, rehman2026autoregressive}. The energy function is evaluated using the Amber14 force field with OBC1 implicit solvent. 

\paragraph{Transition metrics.} To enable a systematic comparison of the transition dynamics across peptides and solvers, we measure the total variation of meta-stable state occupancy $\pi$-TV and transition probabilities $\mathbb{P}$-TV. This requires (1) identifying the meta-stable states and (2) fitting a Markov State Model (MSM) over these discrete states for each trajectory.

To identify the meta-stable states, we compute a time-lagged independent component analysis (TICA) for each peptide sequence using the test set reference trajectory from \textit{Many Peptides} \citep{tan2025manypeptides}. We then use the deeptime package \citep{hoffmann2021deeptime} to cluster the TICA into $K$ meta-stable states. Next, we fit the MSM with a maximum likelihood objective over the identified states for each generated trajectory \citep{hoffmann2021deeptime}.

The MSM can then be queried to obtain the occupation probability for each state $\pi_i$ and transition probabilities between states $i \rightarrow j$ at a lag time $\tau$, denoted by $P_{ij}$. Given a reference $(\hat{\pi}_i, \hat{P}_{ij})$ and predicted $(\pi_i, P_{ij})$, the $\pi$-TV and $\mathbb{P}$-TV metrics are calculated by
\begin{equation}
\begin{aligned}
    \pi\text{-}\mathrm{TV}
    &= \frac{1}{2}\sum_{i=1}^{K}
       \left|\hat{\pi}_i-\pi_i\right|, \\
    \mathbb{P}\text{-}\mathrm{TV}
    &= \frac{1}{2K}\sum_{i=1}^{K}\sum_{j=1}^{K}
       \left|\hat{P}_{ij} -P_{ij}\right|.
\end{aligned}
\end{equation}

\section{Additional Results}
\subsection{Single-molecule dynamics}
\begin{figure}[b]
  \centering
  \includegraphics[width=0.7\textwidth]{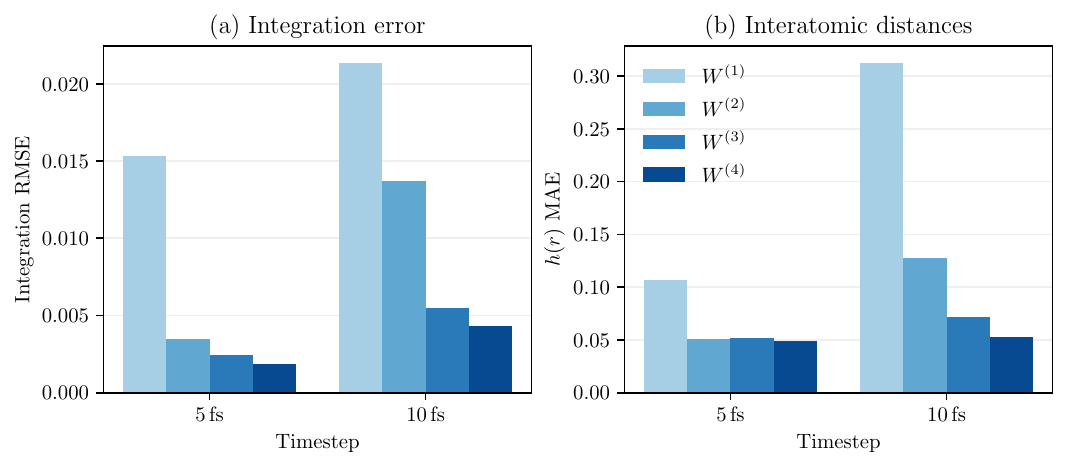}
  \caption{Polynomial degree ablation on the ethanol dataset. The degree is varied over $N=1, 2, 3, 4$ with the polynomial expansion denoted by $W^{(N)}$.}
  \label{fig:ethanol-polynomial-degree-ablation}
\end{figure}

\paragraph{Polynomial degree.} In \cref{fig:ethanol-polynomial-degree-ablation} we ablate the effect of the polynomial degree on the integration RMSE and interatomic distances. We see that as the degree increases the integration RMSE and interatomic distance MAE decrease, with the most pronounced decrease at large timesteps. The effect is most significant when moving from increment-only information $W^{(1)}$ to inclusion of the first higher-order coefficient $W^{(2)}$.

\begin{figure}
  \centering
  \includegraphics[width=0.95\textwidth]{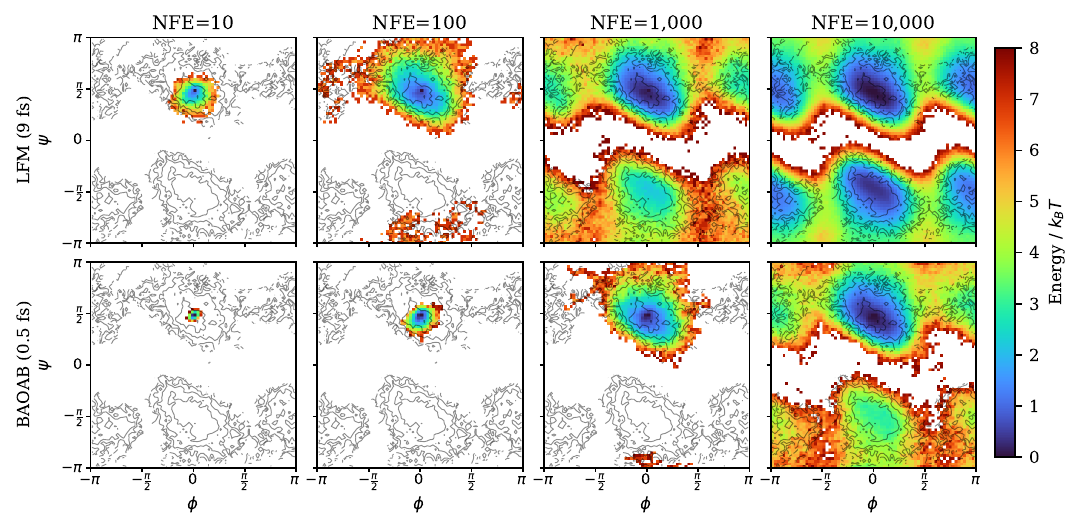}
  \caption{Torsional state-space exploration for the aspirin molecule. The LFM (\emph{top}) and BAOAB (\emph{bottom}) integrators are initialized from a biased start and iterated for $10{,}000$ function evaluations.}
  \label{fig:aspirin-torsions}
\end{figure}

\paragraph{Boltzmann convergence.} In \cref{fig:aspirin-torsions} we plot the Boltzmann convergence dynamics for the aspirin molecule over the state-space defined by the two torsion angles. The LFM ($9$fs) and BAOAB ($0.5$fs) integrators are initialized from a biased start and iterated for $10{,}000$ function evaluations. At matched function evaluations, the state-space coverage obtained by the LFM is significantly increased in comparison to BAOAB, as expected.

\subsection{Transferable dynamics}
\begin{figure}
  \centering
  \includegraphics[width=0.9\textwidth]{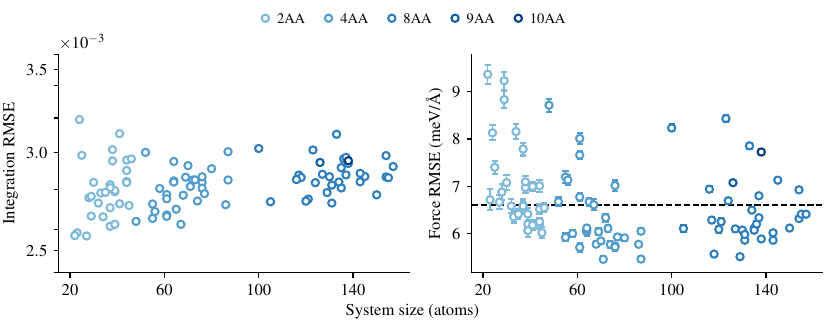}
  \caption{Integration and force errors vs.\ system size on 92 \textit{Many Peptides} test systems. \emph{Left}: integration RMSE at a $5$fs timestep. \emph{Right}: force RMSE evaluated on the $h=0$ force network.}
  \label{fig:transfer-strong-size}
  \label{fig:transfer-force-size}
\end{figure}

\paragraph{Integration and force error.} In \cref{fig:transfer-force-size}, we show the integration RMSE and force RMSE (obtained from the $h=0$ force network) as a function of the system size. As the system size increases, the integration RMSE is approximately constant with a slight increase for larger systems. The force RMSE is also shown to be constant with system size and the mean force RMSE is $6.61$meV/\AA.

\begin{figure}
    \centering
    \includegraphics[width=\linewidth]{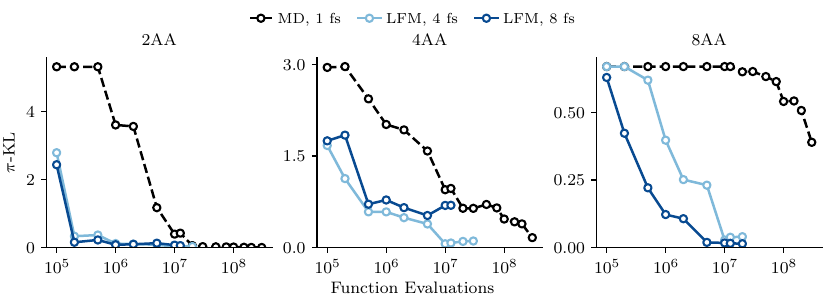}
    \caption{Meta-stable state occupancy Kullback-Leibler (KL) divergence to reference dataset occupancy, $\pi$-KL, in terms of the number of function evaluations.}
    \label{fig:occupancy-compute}
\end{figure}

\paragraph{State occupancy.} As was shown in \cref{tab:lfm-timestep-ensemble}, the meta-stable state occupancy obtained by the large-timestep LFM is improved in comparison to MD at fixed function evaluations. This data is visualized in \cref{fig:occupancy-compute}, where the convergence is shown in terms of the number of function evaluations taken during integration. At matched function evaluations, the state occupancy of the LFM is significantly improved over MD.

\begin{figure}
    \centering
    \includegraphics[width=\linewidth]{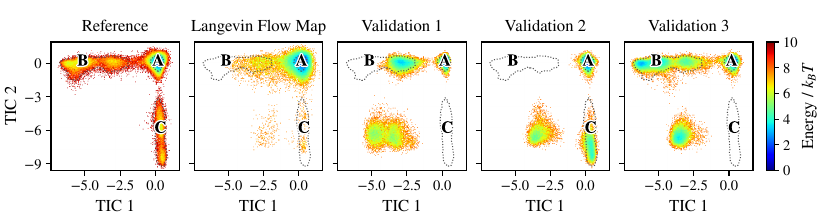}
    \caption{Meta-stable mode not present in reference MD for the ANKSMIEA octapeptide, yet visited by the transferable LFM. Validation MD over three repeats seeded in this state found this mode to be stable for 11.5, 6.8 and 15.8ns respectively.}
    \label{fig:octa-exploration}
\end{figure}

\paragraph{State discovery.} In \cref{fig:octa-exploration}, we show the meta-stable mode discovered by the LFM for the unseen octapeptide ANKSMIEA. This mode is not present in the reference \textit{Many Peptides} test set obtained from long timescale MD. To verify this state is stable, we initialize validation MD trajectories from LFM samples in this state and integrate using the Amber14 force-field. This validation demonstrates that this state is indeed stable, with survival times of 11.5, 6.8 and 15.8ns for each trajectory, respectively.

\begin{figure}
  \centering
  \includegraphics[width=0.6\textwidth]{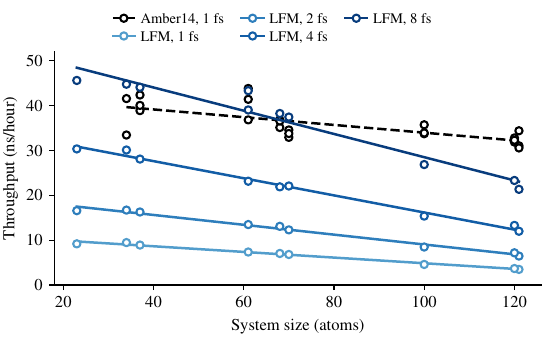}
  \caption{Simulation throughput on 1x RTX PRO 6000 Blackwell. OpenMM Amber14 OBC1 at $1$fs (black, dashed), LFM at $1$--$8$fs (blues).}
  \label{fig:transfer-throughput}
\end{figure}

\paragraph{Integration throughput.} In \cref{fig:transfer-throughput}, we calculate the integration throughput for the transferable LFM and OpenMM Amber14 force-field. At matched timesteps the Amber14 force-field is significantly faster than the LFM, as expected. However, at $8$fs timesteps the LFM obtains comparable simulation throughput to Amber14. This indicates the possibility to obtain MLFFs with comparable or faster integration speeds than classical MD.

\section{Extended Related Work}
\paragraph{Boltzmann Generators.} While this current work focuses on learning the dynamics of molecular systems, Boltzmann Generators (BGs) instead attempt to predict equilibrium molecular structures directly \citep{noe2019boltzmann}. BGs learn a generative model $p_\theta(x)$ that acts as an equilibrium proposal distribution. The samples obtained from this distribution can then be reweighted to the target equilibrium distribution with self-normalized importance sampling. The choice of architecture for $p_\theta(x)$ is an active area of research, focusing on normalizing flows \citep{klein2023equivariant, klein2024transferable, tan2025scalable, rehman2026efficient} and recently autoregressive architectures \citep{rehman2026autoregressive}.

\paragraph{Neural SDEs.} Several works have looked at learning the vector fields of stochastic differential equations from a dataset of reference trajectories with applications to timeseries. These methods differ in their formulation of the objective function; for example via variational inference \citep{li2020scalable}, generative adversarial networks \citep{kidger2021neural}, signature kernel scores \citep{issa2023non} and maximum likelihood \citep{dietrich2023learning}. These methods all require reference trajectories during training.

\paragraph{Machine-learned force fields.} In molecular dynamics, the ground truth Langevin dynamics vector fields can be queried directly and therefore neither trajectories or training integration is required. The force-field can be regressed against directly, leading to the literature on machine-learned force-fields \citep{behler2007generalized, rupp2012fast, chmiela2018towards, noe2020machine, keith2021combining, unke2021machine}.

\paragraph{Stochastic Flow Maps.} In the application of generative modelling, recent work has proposed methods to learn the solution map of a neural differential equation directly. These methods were first developed for deterministic ODE models \citep{boffi2024flow, frans2025one, geng2026mean, boffi2026build, ripken2026learning}. More recently, these methods have been extended to obtain weak (distributional) solutions to stochastic dynamics \citep{kiyohara2026neural, potaptchik2026meta, holderrieth2026diamond} but require explicit trajectory supervision, and strong (pathwise) solutions \citep{mccallum2026strong} which can be trained without ordered trajectory data. In the case of diffusion models, the trajectory supervision required by weak formulations can be obtained without integration by querying the forward diffusion process or interpolant.

\paragraph{Transition Path Sampling.} Recent literature has developed methods for obtaining the transition pathways of a system more efficiently via reinforcement learning \citep{das2021reinforcement, rose2021reinforcement}, differentiable simulation \citep{sipka2023differentiable}, Doob's $h$-transform learning \citep{singh2023variational, du2024doob} and Onsager-Machlup action minimization \citep{raja2025action}. These approaches require stochastic sampling procedures and therefore can be paired with Langevin Flow Maps to obtain more efficient algorithms. We hope future work may explore this combination of algorithms.

\section{Limitations \& future work}
This work introduces an objective to learn the stochastic solution map of Langevin Dynamics. The resulting model class obtained by optimizing this objective, called Langevin Flow Maps, can be used to run large-timestep molecular dynamics and efficiently sample transition pathways. The current limitations of this approach and directions for future work are discussed below.

\paragraph{Parameterization.} The current Euler-Maruyama-like update step aims to not constrain the architecture to any specific small-timestep solver or scheme. This allows the network a large amount of freedom during optimization. However, this parameterization does not constrain certain symmetries of the update rule such as equipartition or reversibility. It may be advantageous to introduce these constraints into the update rule; this line of argument can be explored in future work.

\paragraph{Architecture.} The current model architecture uses a translation invariant transformer and direct force prediction. To ensure translation invariance, the positional embeddings are computed as pair-wise distances, $r_{ij}$. This embedding can become expensive at large system sizes since the input computation scales as $O(N^2)$, for an $N$-atom system. The MLFF literature has approached this cost by separating local and global computation \citep{unke2021spookynet}. 

The direct force prediction head provided improved performance in early experiments over a conservative energy prediction architecture. However, it is possible to parameterize a hybrid prediction head with $f_{s, t}^\theta(x_s, v_s, I_{s, t}^N) = -\nabla U^\theta(x_s) + (t-s)\bar{f}_{s, t}^\theta(x_s, v_s, I_{s, t}^N)$. This retains conservative forces at $s=t$ but allows the network the freedom of direct prediction for $t>s$. These higher-order flow map architectures could be explored in future work.

\paragraph{Transferability.} The transferable LFM model demonstrated that it is possible to learn a general integrator for peptide systems that can predict accurate dynamics and transition pathways for unseen peptides. However, the model's predicted Boltzmann ensemble for unseen systems was shown to be biased at large timesteps as measured by the energy and dihedral angles, despite accurate force field prediction as $h\rightarrow 0$. This suggests that future work should focus on improving the transferable integration performance such that, as suggested by throughput analysis, the model can obtain a speedup over classical molecular dynamics simulation.

\end{document}

%% file: iclr2027/math_commands.tex
\usepackage{amsmath,amsfonts,bm}
\usepackage{xspace}

\def\eqref#1{equation~\ref{#1}}

\def\1{\bm{1}}

\DeclareMathAlphabet{\mathsfit}{\encodingdefault}{\sfdefault}{m}{sl}
\SetMathAlphabet{\mathsfit}{bold}{\encodingdefault}{\sfdefault}{bx}{n}



%% file: figures/sampling_table.tex
\begingroup
\fontsize{7.5}{9}\selectfont
\setlength{\tabcolsep}{2pt}
\begin{tabular}{@{}lllr@{\hspace{6pt}}r@{\hspace{6pt}}r@{\hspace{6pt}}r@{\hspace{10pt}}r@{\hspace{6pt}}r@{\hspace{6pt}}r@{\hspace{6pt}}r@{\hspace{10pt}}r@{\hspace{6pt}}r@{\hspace{6pt}}r@{\hspace{6pt}}r@{}}
\toprule
 & & & \multicolumn{4}{c}{2AA} & \multicolumn{4}{c}{4AA} & \multicolumn{4}{c}{8AA} \\
\cmidrule(lr){4-7}\cmidrule(lr){8-11}\cmidrule(lr){12-15}
Forces & Integrator & Timestep & $\mathcal{E}$-$W_2$ & $\mathbb{T}$-$W_2$ & $\pi$-TV & $\mathbb{P}$-TV & $\mathcal{E}$-$W_2$ & $\mathbb{T}$-$W_2$ & $\pi$-TV & $\mathbb{P}$-TV & $\mathcal{E}$-$W_2$ & $\mathbb{T}$-$W_2$ & $\pi$-TV & $\mathbb{P}$-TV \\
\midrule
LFM & LFM & 8~fs & 3.64 & 0.81 & 0.13 & 0.10 & 11.75 & 2.31 & 0.26 & 0.19 & 25.91 & 4.17 & \textbf{0.03} & \textbf{0.45} \\
LFM & LFM & 4~fs & 2.73 & 0.61 & \textbf{0.07} & \textbf{0.04} & 8.76 & 2.10 & \textbf{0.08} & \textbf{0.18} & 20.90 & 4.08 & 0.03 & 0.49 \\
LFM & LFM & 2~fs & 1.81 & 0.62 & 0.28 & 0.06 & 7.45 & 1.95 & 0.14 & 0.32 & 18.57 & 4.27 & 0.19 & 0.48 \\
LFM & LFM & 1~fs & 1.20 & 0.73 & 0.21 & 0.07 & 5.08 & 1.70 & 0.15 & 0.43 & 11.45 & 4.26 & 0.05 & 0.47 \\
\midrule
LFM & BAOAB & 1~fs & 0.49 & 0.44 & 0.21 & 0.13 & 1.22 & 1.42 & 0.34 & 0.31 & 6.10 & \textbf{3.18} & 0.07 & 0.49 \\
\midrule
Amber14 & BAOAB & 1~fs & \textbf{0.30} & \textbf{0.35} & 0.15 & 0.12 & \textbf{1.04} & \textbf{1.13} & 0.18 & 0.27 & \textbf{4.04} & 3.41 & 0.07 & 0.47 \\
\bottomrule
\end{tabular}
\endgroup

%% file: references.bib
@article{mccallum2026strong,
  title={Strong stochastic flow maps},
  author={McCallum, Sam and Blasingame, Zander W and Herschell, Timothy and Rindtorff, Niklas and Tong, Alexander and Foster, James},
  journal={arXiv preprint arXiv:2606.01086},
  year={2026}
}

@article{leimkuhler2013rational,
  title={Rational construction of stochastic numerical methods for molecular sampling},
  author={Leimkuhler, Benedict and Matthews, Charles},
  journal={Applied Mathematics Research eXpress},
  volume={2013},
  number={1},
  pages={34--56},
  year={2013},
  publisher={Oxford University Press}
}

@article{ripken2026learning,
  title={Learning {Hamiltonian} Flow Maps: Mean Flow Consistency for Large-Timestep Molecular Dynamics},
  author={Ripken, Winfried and Plainer, Michael and Lied, Gregor and Frank, Thorben and Unke, Oliver T and Chmiela, Stefan and No{\'e}, Frank and M{\"u}ller, Klaus-Robert},
  journal={International Conference on Machine Learning},
  year={2026}
}

@article{boffi2026build,
  title={How to build a consistency model: Learning flow maps via self-distillation},
  author={Boffi, Nicholas and Albergo, Michael and Vanden-Eijnden, Eric},
  journal={Advances in Neural Information Processing Systems},
  volume={38},
  pages={33346--33382},
  year={2025}
}

@article{fu2022forces,
  title={Forces are not enough: Benchmark and critical evaluation for machine learning force fields with molecular simulations},
  author={Fu, Xiang and Wu, Zhenghao and Wang, Wujie and Xie, Tian and Keten, Sinan and Gomez-Bombarelli, Rafael and Jaakkola, Tommi},
  journal={Transactions on Machine Learning Research},
  year={2023}
}

@article{kohler2021smooth,
  title={Smooth normalizing flows},
  author={K{\"o}hler, Jonas and Kr{\"a}mer, Andreas and No{\'e}, Frank},
  journal={Advances in Neural Information Processing Systems},
  volume={34},
  pages={2796--2809},
  year={2021}
}

@article{foster2020optimal,
  title={An optimal polynomial approximation of {Brownian} motion},
  author={Foster, James and Lyons, Terry and Oberhauser, Harald},
  journal={SIAM Journal on Numerical Analysis},
  volume={58},
  number={3},
  pages={1393--1421},
  year={2020},
  publisher={SIAM}
}

@article{habermann2021semicircle,
  title={A semicircle law and decorrelation phenomena for iterated {Kolmogorov} loops},
  author={Habermann, Karen},
  journal={Journal of the London Mathematical Society},
  volume={103},
  number={2},
  pages={558--586},
  year={2021},
}

@article{tan2025manypeptides,
 author = {Tan, Charlie and Hassan, Majdi and Klein, Leon and Syed, Saifuddin and Beaini, Dominique and Bronstein, Michael and Tong, Alexander and Neklyudov, Kirill},
 journal = {Advances in Neural Information Processing Systems},
 pages = {94290--94325},
 title = {Amortized Sampling with Transferable Normalizing Flows},
 volume = {38},
 year = {2025}
}

@article{hollingsworth2018molecular,
  title={Molecular dynamics simulation for all},
  author={Hollingsworth, Scott A and Dror, Ron O},
  journal={Neuron},
  volume={99},
  number={6},
  pages={1129--1143},
  year={2018},
  publisher={Elsevier}
}

@book{goldstein2011classical,
  title={Classical Mechanics},
  author={Goldstein, Herbert and Poole, Charles P. and Safko, John},
  edition={3},
  year={2011},
  publisher={Pearson}
}

@article{tuckerman2000understanding,
  title={Understanding modern molecular dynamics: Techniques and applications},
  author={Tuckerman, Mark E and Martyna, Glenn J},
  journal={The Journal of Physical Chemistry B},
  volume={104},
  number={2},
  pages={159--178},
  year={2000},
  publisher={ACS Publications}
}

@book{frenkel1997understanding,
  title={Understanding molecular simulation: from algorithms to applications},
  author={Frenkel, Daan and Smit, Berend},
  year={1996},
  publisher={Academic Press}
}

@article{noe2020machine,
  title={Machine learning for molecular simulation},
  author={No{\'e}, Frank and Tkatchenko, Alexandre and M{\"u}ller, Klaus-Robert and Clementi, Cecilia},
  journal={Annual Review of Physical Chemistry},
  volume={71},
  number={1},
  pages={361--390},
  year={2020},
  publisher={Annual Reviews}
}

@article{keith2021combining,
  title={Combining machine learning and computational chemistry for predictive insights into chemical systems},
  author={Keith, John A and Vassilev-Galindo, Valentin and Cheng, Bingqing and Chmiela, Stefan and Gastegger, Michael and M{\"u}ller, Klaus-Robert and Tkatchenko, Alexandre},
  journal={Chemical Reviews},
  volume={121},
  number={16},
  pages={9816--9872},
  year={2021},
  publisher={ACS Publications}
}

@article{unke2021machine,
  title={Machine learning force fields},
  author={Unke, Oliver T and Chmiela, Stefan and Sauceda, Huziel E and Gastegger, Michael and Poltavsky, Igor and Sch{\"u}tt, Kristof T and Tkatchenko, Alexandre and M{\"u}uller, Klaus-Robert},
  journal={Chemical Reviews},
  volume={121},
  number={16},
  pages={10142--10186},
  year={2021},
  publisher={ACS Publications}
}

@article{wang2025design,
  title={On the design space between molecular mechanics and machine learning force fields},
  author={Wang, Yuanqing and Takaba, Kenichiro and Chen, Michael S and Wieder, Marcus and Xu, Yuzhi and Zhu, Tong and Zhang, John ZH and Nagle, Arnav and Yu, Kuang and Wang, Xinyan and others},
  journal={Applied Physics Reviews},
  volume={12},
  number={2},
  year={2025},
  publisher={AIP Publishing}
}

@article{klein2023timewarp,
  title={Timewarp: Transferable acceleration of molecular dynamics by learning time-coarsened dynamics},
  author={Klein, Leon and Foong, Andrew and Fjelde, Tor and Mlodozeniec, Bruno and Brockschmidt, Marc and Nowozin, Sebastian and No{\'e}, Frank and Tomioka, Ryota},
  journal={Advances in Neural Information Processing Systems},
  volume={36},
  pages={52863--52883},
  year={2023}
}

@article{schreiner2023implicit,
  title={Implicit transfer operator learning: Multiple time-resolution models for molecular dynamics},
  author={Schreiner, Mathias and Winther, Ole and Olsson, Simon},
  journal={Advances in Neural Information Processing Systems},
  volume={36},
  pages={36449--36462},
  year={2023}
}

@article{diez2026transferable,
  title={Transferable generative models bridge femtosecond to nanosecond time-step molecular dynamics},
  author={Diez, Juan Viguera and Schreiner, Mathias and Olsson, Simon},
  journal={Science Advances},
  volume={12},
  number={15},
  pages={eaed2333},
  year={2026},
  publisher={American Association for the Advancement of Science}
}

@article{geng2026mean,
  title={Mean flows for one-step generative modeling},
  author={Geng, Zhengyang and Deng, Mingyang and Bai, Xingjian and Kolter, Zico and He, Kaiming},
  journal={Advances in Neural Information Processing Systems},
  volume={38},
  pages={75460--75482},
  year={2025}
}

@article{pastor1988analysis,
  title={An analysis of the accuracy of {Langevin} and molecular dynamics algorithms},
  author={Pastor, Richard W and Brooks, Bernard R and Szabo, Attila},
  journal={Molecular Physics},
  volume={65},
  number={6},
  pages={1409--1419},
  year={1988},
  publisher={Taylor \& Francis}
}

@inproceedings{frans2025one,
  title={One step diffusion via shortcut models},
  author={Frans, Kevin and Hafner, Danijar and Levine, Sergey and Abbeel, Pieter},
  booktitle={International Conference on Learning Representations},
  volume={2025},
  pages={34668--34684},
  year={2025}
}

@article{chmiela2017machine,
  title={Machine learning of accurate energy-conserving molecular force fields},
  author={Chmiela, Stefan and Tkatchenko, Alexandre and Sauceda, Huziel E and Poltavsky, Igor and Sch{\"u}tt, Kristof T and M{\"u}ller, Klaus-Robert},
  journal={Science Advances},
  volume={3},
  number={5},
  pages={e1603015},
  year={2017},
  publisher={American Association for the Advancement of Science}
}

@article{noe2019boltzmann,
  title={{Boltzmann} generators: Sampling equilibrium states of many-body systems with deep learning},
  author={No{\'e}, Frank and Olsson, Simon and K{\"o}hler, Jonas and Wu, Hao},
  journal={Science},
  volume={365},
  number={6457},
  pages={eaaw1147},
  year={2019},
  publisher={American Association for the Advancement of Science}
}

@article{tan2025scalable,
  title={Scalable equilibrium sampling with sequential {Boltzmann} generators},
  author={Tan, Charlie B and Bose, Avishek Joey and Lin, Chen and Klein, Leon and Bronstein, Michael M and Tong, Alexander},
  journal={International Conference on Machine Learning},
  year={2025}
}

@inproceedings{rehman2026efficient,
  title={Efficient regression-based training of normalizing flows for {Boltzmann} generators},
  author={Rehman, Danyal and Davis, Oscar and Lu, Jiarui and Tang, Jian and Bronstein, Michael and Bengio, Yoshua and Tong, Alexander and Bose, Joey},
  booktitle={International Conference on Learning Representations},
  volume={2026},
  pages={127171--127194},
  year={2026}
}

@article{klein2023equivariant,
  title={Equivariant flow matching},
  author={Klein, Leon and Kr{\"a}mer, Andreas and No{\'e}, Frank},
  journal={Advances in Neural Information Processing Systems},
  volume={36},
  pages={59886--59910},
  year={2023}
}

@article{klein2024transferable,
  title={Transferable boltzmann generators},
  author={Klein, Leon and No{\'e}, Frank},
  journal={Advances in Neural Information Processing Systems},
  volume={37},
  pages={45281--45314},
  year={2024}
}

@article{rehman2026autoregressive,
  title={Autoregressive {Boltzmann} Generators},
  author={Rehman, Danyal and Tan, Charlie B and Bengio, Yoshua and Bose, Avishek Joey and Tong, Alexander},
  journal={International Conference on Machine Learning},
  year={2026}
}

@inproceedings{li2020scalable,
  title={Scalable gradients for stochastic differential equations},
  author={Li, Xuechen and Wong, Ting-Kam Leonard and Chen, Ricky TQ and Duvenaud, David},
  booktitle={International conference on artificial intelligence and statistics},
  pages={3870--3882},
  year={2020},
  organization={PMLR}
}

@inproceedings{kidger2021neural,
  title={Neural {SDEs} as infinite-dimensional {GANs}},
  author={Kidger, Patrick and Foster, James and Li, Xuechen and Oberhauser, Harald and Lyons, Terry J},
  booktitle={International conference on machine learning},
  pages={5453--5463},
  year={2021},
  organization={PMLR}
}

@article{issa2023non,
  title={Non-adversarial training of Neural SDEs with signature kernel scores},
  author={Issa, Zacharia and Horvath, Blanka and Lemercier, Maud and Salvi, Cristopher},
  journal={Advances in Neural Information Processing Systems},
  volume={36},
  pages={11102--11126},
  year={2023}
}

@article{dietrich2023learning,
  title={Learning effective stochastic differential equations from microscopic simulations: Linking stochastic numerics to deep learning},
  author={Dietrich, Felix and Makeev, Alexei and Kevrekidis, George and Evangelou, Nikolaos and Bertalan, Tom and Reich, Sebastian and Kevrekidis, Ioannis G},
  journal={Chaos: An Interdisciplinary Journal of Nonlinear Science},
  volume={33},
  number={2},
  year={2023},
  publisher={AIP Publishing}
}

@article{behler2007generalized,
  title={Generalized neural-network representation of high-dimensional potential-energy surfaces},
  author={Behler, J{\"o}rg and Parrinello, Michele},
  journal={Physical Review Letters},
  volume={98},
  number={14},
  pages={146401},
  year={2007},
  publisher={APS}
}

@article{rupp2012fast,
  title={Fast and accurate modeling of molecular atomization energies with machine learning},
  author={Rupp, Matthias and Tkatchenko, Alexandre and M{\"u}ller, Klaus-Robert and Von Lilienfeld, O Anatole},
  journal={Physical Review Letters},
  volume={108},
  number={5},
  pages={058301},
  year={2012},
  publisher={APS}
}

@article{chmiela2018towards,
  title={Towards exact molecular dynamics simulations with machine-learned force fields},
  author={Chmiela, Stefan and Sauceda, Huziel E and M{\"u}ller, Klaus-Robert and Tkatchenko, Alexandre},
  journal={Nature Communications},
  volume={9},
  number={1},
  pages={3887},
  year={2018},
  publisher={Nature Publishing Group UK London}
}

@article{boffi2024flow,
  title={Flow map matching with stochastic interpolants: A mathematical framework for consistency models},
  author={Boffi, Nicholas M and Albergo, Michael S and Vanden-Eijnden, Eric},
  journal={Transactions on Machine Learning Research},
  year={2025}
}

@article{holderrieth2026diamond,
  title={Diamond maps: Efficient reward alignment via stochastic flow maps},
  author={Holderrieth, Peter and Chen, Douglas and Eyring, Luca and Shah, Ishin and Anantharaman, Giri and He, Yutong and Akata, Zeynep and Jaakkola, Tommi and Boffi, Nicholas Matthew and Simchowitz, Max},
  journal={International Conference on Machine Learning},
  year={2026}
}

@article{kiyohara2026neural,
  title={Neural stochastic flows: Solver-free modelling and inference for {SDE} solutions},
  author={Kiyohara, Naoki and Johns, Edward and Li, Yingzhen},
  journal={Advances in Neural Information Processing Systems},
  volume={38},
  pages={16917--16962},
  year={2025}
}

@article{potaptchik2026meta,
  title={Meta flow maps enable scalable reward alignment},
  author={Potaptchik, Peter and Saravanan, Adhi and Mammadov, Abbas and Prat, Alvaro and Albergo, Michael S and Teh, Yee Whye},
  journal={arXiv preprint arXiv:2601.14430},
  year={2026}
}

@article{thiemann2026force,
  title={Force-free molecular dynamics through autoregressive equivariant networks},
  author={Thiemann, Fabian L and Resch{\"u}tzegger, Thiago and Esposito, Massimiliano and Taddese, Tseden and Olarte-Plata, Juan D and Martelli, Fausto},
  journal={Nature Machine Intelligence},
  pages={1--13},
  year={2026},
  publisher={Nature Publishing Group UK London}
}

@article{bigi2026flashmd,
  title={{FlashMD}: long-stride, universal prediction of molecular dynamics},
  author={Bigi, Filippo and Chong, Sanggyu and Kristiadi, Agustinus and Ceriotti, Michele},
  journal={Advances in Neural Information Processing Systems},
  volume={38},
  pages={88537--88574},
  year={2025}
}

@article{hoffmann2021deeptime,
  title={Deeptime: a Python library for machine learning dynamical models from time series data},
  author={Hoffmann, Moritz and Scherer, Martin Konrad and Hempel, Tim and Mardt, Andreas and de Silva, Brian and Husic, Brooke Elena and Klus, Stefan and Wu, Hao and Kutz, J Nathan and Brunton, Steven and Noé, Frank},
  journal={Machine Learning: Science and Technology},
  year={2021},
  publisher={IOP Publishing}
}

@article{das2021reinforcement,
  title={Reinforcement learning of rare diffusive dynamics},
  author={Das, Avishek and Rose, Dominic C and Garrahan, Juan P and Limmer, David T},
  journal={The Journal of Chemical Physics},
  volume={155},
  number={13},
  year={2021},
  publisher={AIP Publishing}
}

@article{rose2021reinforcement,
  title={A reinforcement learning approach to rare trajectory sampling},
  author={Rose, Dominic C and Mair, Jamie F and Garrahan, Juan P},
  journal={New Journal of Physics},
  volume={23},
  number={1},
  pages={013013},
  year={2021},
  publisher={IOP Publishing}
}

@inproceedings{sipka2023differentiable,
  title={Differentiable simulations for enhanced sampling of rare events},
  author={Sipka, Martin and Dietschreit, Johannes CB and Grajciar, Luk{\'a}{\v{s}} and G{\'o}mez-Bombarelli, Rafael},
  booktitle={International Conference on Machine Learning},
  pages={31990--32007},
  year={2023},
  organization={PMLR}
}

@article{singh2023variational,
  title={Variational deep learning of equilibrium transition path ensembles},
  author={Singh, Aditya N and Limmer, David T},
  journal={The Journal of Chemical Physics},
  volume={159},
  number={2},
  year={2023},
  publisher={AIP Publishing}
}

@article{du2024doob,
  title={Doob's lagrangian: A sample-efficient variational approach to transition path sampling},
  author={Du, Yuanqi and Plainer, Michael and Brekelmans, Rob and Duan, Chenru and Noe, Frank and Gomes, Carla P and Aspuru-Guzik, Alan and Neklyudov, Kirill},
  journal={Advances in Neural Information Processing Systems},
  volume={37},
  pages={65791--65822},
  year={2024}
}

@article{raja2025action,
  title={Action-minimization meets generative modeling: Efficient transition path sampling with the onsager-machlup functional},
  author={Raja, Sanjeev and {\v{S}}{\'\i}pka, Martin and Psenka, Michael and Kreiman, Tobias and Pavelka, Michal and Krishnapriyan, Aditi S},
  journal={International Conference on Machine Learning},
  year={2025}
}

@article{unke2021spookynet,
  title={SpookyNet: Learning force fields with electronic degrees of freedom and nonlocal effects},
  author={Unke, Oliver T and Chmiela, Stefan and Gastegger, Michael and Sch{\"u}tt, Kristof T and Sauceda, Huziel E and M{\"u}ller, Klaus-Robert},
  journal={Nature communications},
  volume={12},
  number={1},
  pages={7273},
  year={2021},
  publisher={Nature Publishing Group UK London}
}
